\documentclass{article}

\PassOptionsToPackage{numbers, compress}{natbib}

\usepackage[preprint]{neurips_2024}

\usepackage[utf8]{inputenc} 
\usepackage[T1]{fontenc}    
\usepackage{hyperref}       
\usepackage{url}            
\usepackage{booktabs}       
\usepackage{amsfonts}       
\usepackage{nicefrac}       
\usepackage{microtype}      
\usepackage{xcolor}         
\usepackage{amsmath}
\usepackage{graphicx} 
\usepackage{wrapfig}
\usepackage{xcolor}
\usepackage{algorithm}
\usepackage{algpseudocode}
\usepackage{comment}
\definecolor{prompt_gray}{RGB}{220,220,220}

\title{Toward Conversational Agents with Context and Time Sensitive Long-term Memory}

\author{%
  Nick Alonso\\
  Zyphra\\
  \texttt{nick@zyphra.com} \\
  \And
  Tomás Figliolia \\
  Zyphra \\
  \texttt{tom@zyphra.com} \\
  \AND
  Anthony Ndirango \\
  Zyphra \\
  \texttt{anthony@zyphra.com} \\
  \And
  Beren Millidge \\
  Zyphra \\
  \texttt{beren@zyphra.com} \\
}

\begin{document}

\maketitle

\begin{abstract}
There has recently been growing interest in conversational agents with long-term memory which has led to the rapid development of language models that use retrieval-augmented generation (RAG). Until recently, most work on RAG has focused on information retrieval from large databases of texts, like Wikipedia, rather than information from long-form conversations. In this paper, we argue that effective retrieval from long-form conversational data faces two unique problems compared to static database retrieval: 1) time/event-based queries, which requires the model to retrieve information about previous conversations based on time or the order of a conversational event (e.g., the third conversation on Tuesday), and 2) ambiguous queries that require surrounding conversational context to understand. To better develop RAG-based agents that can deal with these challenges, we generate a new dataset of ambiguous and time-based questions that build upon a recent dataset of long-form, simulated conversations, and demonstrate that standard RAG based approaches handle such questions poorly. We then develop a novel retrieval model which combines chained-of-table search methods, standard vector-database retrieval, and a prompting method to disambiguate queries, and demonstrate that this approach substantially improves over current methods at solving these tasks. We believe that this new dataset and more advanced RAG agent can act as a key benchmark and stepping stone towards effective memory augmented conversational agents that can be used in a wide variety of AI applications.\footnote{Dataset can be found at this link: \url{https://github.com/Zyphra/TemporalMemoryDataset.git}}
\end{abstract}

\section{Introduction}

Conversational agents, such as chatbots, personal assistants, and language interfaced operating systems, are currently seeing rapid development and interest both in academia and industry (e.g., \cite{jang2023conversation, maharana2024evaluating, zhong2024memorybank, lee2023prompted, lu2023memochat, wu2023brief}). One specific area of interest is in conversational agents that utilize retrieval-augmented generation (RAG) to imbue these agents with long-term memory. However, popular QA benchmarks that typically test RAG systems, focus primarily on information retrieval from a static database of texts, such as Wikipedia (e.g., \cite{kwiatkowski2019natural,yang2018hotpotqa}. However, the increasing importance of conversational agents raises the question of how to address the unique challenges RAG models face in conversational contexts that they do not in offline, database retrieval contexts. In this paper, we focus on what we see as two crucial challenges conversational agents face that are not tested in most standard database retrieval benchmarks:
\begin{enumerate}
  \item \textbf{Conversational Meta-Data Based Queries.} In conversational contexts, a common sort of query refers to meta-data (e.g. time, date, or speaker) associated with previous conversations. For example, one could plausibly ask "what were we discussing yesterday morning, again?", "what was that idea we were working on last time?", or "summarize what Jason talked about in our meeting from January 6th.". These questions are not specifying what was talked about, but are instead asking the model to specify what was talked about given some meta-data (such as time) associated with a conversational event. Such questions point to a class of common questions an conversational agent may face, which cannot be answered without some ability to retrieve information about previous conversations based on conversational meta-data, rather than semantic retrieval alone.
  \item \textbf{Ambiguous Questions. } In conversation, it is normal to speak with pronouns (he, she, it, they, etc.) and demonstratives ('this', 'that', etc.), which are ambiguous without an understanding of preceding conversational context. Although understanding this context is trivial for generation by LLMs, such statements will fool naive RAG systems as we discuss below.
\end{enumerate}

There is currently a lack of good benchmarks and models that explicitly and directly address these challenges for retrieval systems for conversational agents. Although, there exists a QA benchmark that tests ambiguous questions \cite{dalton2020trec, dalton2021cast}, it does not test conversational agents that are retrieving from records of chat history and does not test ambiguous queries that refer to meta-data rather than text content. Further, recent work has created benchmarks which test long-term memory in conversational agents (e.g., \cite{jang2023conversation, maharana2024evaluating, goodai_benchmark2023}) that do not directly, or deeply test meta-data retrieval or ambiguous questions. With respect to models, although IR models have long existed for searching through databases with meta-data, such as IR models for tabular data, such models are not standard in conversational agents that use RAG (e.g., see \cite{jang2023conversation, maharana2024evaluating, zhong2024memorybank, lee2023prompted,goodai_benchmark2023, lu2023memochat}), which primarily utilize semantic, vector database search.

There is, therefore, a need for datasets and benchmarks that directly test a conversational agent's ability to recall information based on queries which are ambiguous and which refer to conversational meta-data. In this paper, we attempt to make progress in this direction through two main contributions. First, we construct a dataset and benchmark for conversational agents that directly tests these two capabilities. We build upon an existing dataset of long-form dialogues to generate a dataset of questions that 1) refer to conversational meta-data, 2) ambiguous questions that refer to conversational meta-data, and 3) questions that refer to a combination of meta-data and conversational content. Second, we develop a novel retrieval model that combines standard vector database search with a tabular search method known as chain-of-tables \cite{wang2024chain} for retrieving from chat logs using meta-data based queries, and we improve the performance of this model through the use of a classifier that decides whether the query is referencing meta-data, content, or both. We find that this system significantly outperforms existing retrieval systems in the literature, especially w.r.t. the common problems we have identified. Ultimately, we believe that our work can act as a stepping stone toward more intelligent conversational agents with long-term memory.

\section{Related Works}\label{sec:rel_work}

\textbf{Long-Context LLMs} Significant recent attention has also been paid to 'needle in the haystack' tasks for LLMs with long contexts, where the task requires reporting information stored in a small portion of text present in long, usually > 100,000, context of an LLM (e.g., \cite{ding2023longnet, chen2023extending, bertsch2024unlimiformer, bulatov2023scaling, ma2024megalodon}). Long-context models often have difficulty finding and attending to relevant tokens while ignoring all the other information stored in context. We are similarly interested in how a model may retrieve information contained in a small portion of text that exists in a long stream of input text. However, contrary to this line of work, we are 1) interested in computationally cheap, affordable ways of performing this task that does not require long context models, and 2) interested in search for a small chunk of text based on its meta-data (e.g., date or time) rather than on content contained in its text.

\textbf{Retrieval Augmented Generation} RAG models \cite{lewis2020retrieval} work by storing a database of text chunks, each of which is associated with a semantic embedding vector, which is then used as a key during vector database retrieval. Retrieved text can be injected directly into the context or injected at hidden layers of LLMs \cite{borgeaud2022improving} RAG models often show improved performance, both in terms of perplexity and QA accuracy, over LLMs that do not utilize RAG but have the same or more overall size (e.g., \cite{lewis2020retrieval, borgeaud2022improving}. RAG models are often used on QA benchmarks that use a standard database of texts like Wikipedia (e.g., \cite{kwiatkowski2019natural, yang2018hotpotqa}. More recently, RAG applied to streams of conversational responses has been investigated, which we now discuss.

\textbf{Long-term Dialogue.} There has been a growing interest in testing information retrieval (IR) in conversational agents (e.g., \cite{jang2023conversation, maharana2024evaluating, zhong2024memorybank, lee2023prompted, lu2023memochat, wu2023brief}). Unlike benchmarks that test IR on static datasets, such as Wikipedia (e.g., \cite{kwiatkowski2019natural, yang2018hotpotqa}), these benchmarks seek to test a conversational agent's ability to recall and report information that may have been communicated by the user far in the past of the conversation history (i.e. out of immediate context of the model). Until very recently, however most datasets of conversation logs were relatively short, involving only a few sessions/conversations between an agent and a single user (e.g., \cite{jang2023conversation, zhong2024memorybank, lee2023prompted, lu2023memochat, wu2023brief}). More recently, there have been efforts to create benchmarks that involve much longer dialogues to better test long-term memory abilities vs within context retrieval. The company GoodAI recently released a series of benchmarks that test a conversational agent's ability to retrieve information that may have been communicated tens or hundreds of thousands of tokens in the past \cite{goodai_benchmark2023}. This test works by interleaving simulated small talk and filler text with questions from the user. The recently developed LoCoMO dataset \cite{maharana2024evaluating} is a high-quality dataset of long-form conversations between two simulated, GPT4-based agents that involve many sessions worth of conversation. The LoCoMo benchmark uses questions that involve summarization of past portions of conversation, one-hop and multi-hop question answering, and temporal reasoning. Both the GoodAI and LoCoMo benchmarks test IR in conversational agents in many useful ways, but do not directly test or focus on ambiguous question answering or meta-data-based IR.

\begin{figure}[t]
\includegraphics[width=.98\textwidth]{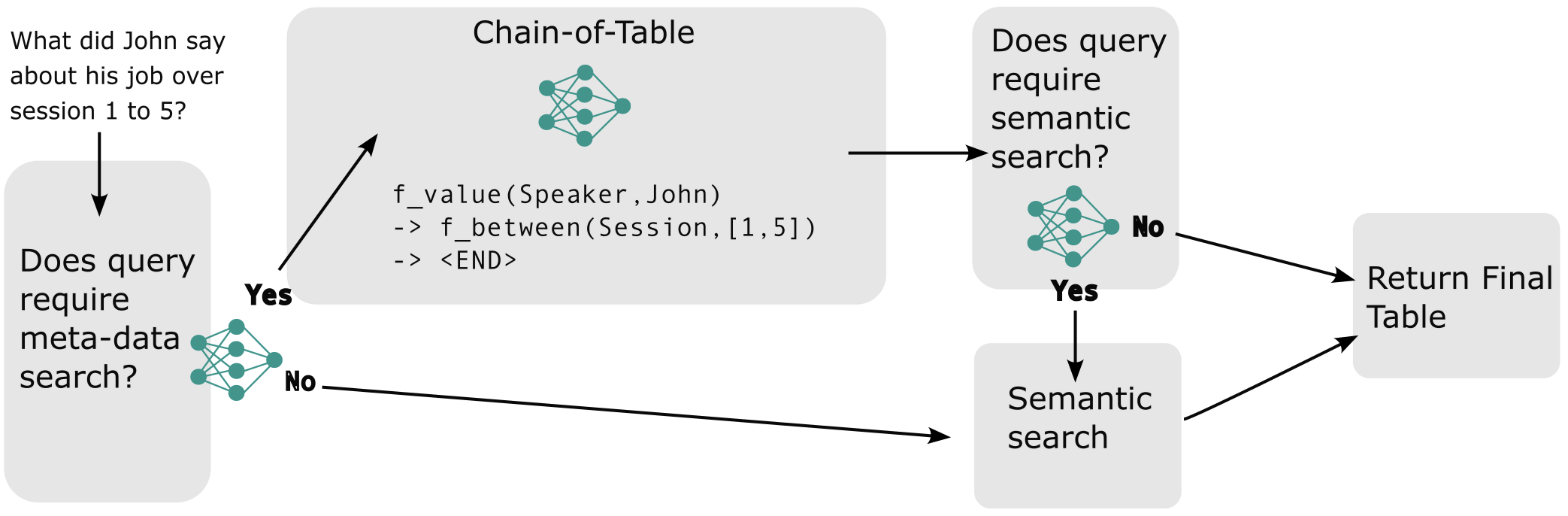}
\centering
\caption{Depiction of our combined tabular and semantic vector-search method.}
\end{figure}

\textbf{Temporal Reasoning in LLMs.} Although, the LoCoMo dataset \cite{maharana2024evaluating} tests temporal reasoning, this reasoning only requires the retrieval of items using content-based retrieval, e.g., the question 'how long did it take for Greg to write his novel?', only requires a semantic retrieval system to retrieve statements about novel writing along with their associated timestamps. Other tests of temporal reasoning in LLMs, such as the benchmark by \cite{wang2023tram}, focus on LLMs ability to reason about the frequency, ordering, duration, causality, temporal differences, and other related temporal properties, e.g., multiple choice questions like "Form the following events in chronological order." or "Which of the following events is the longest?". These questions do not specifically require retrieval items from a memory bank based on their temporal properties (e.g., when they occurred) since the questions are usually one-off questions that provide the relevant events names/descriptions within the question.

\textbf{Ambiguous and Conversational Querying.} Common QA benchmarks for language models, like Natural questions \cite{kwiatkowski2019natural} or HotPotQA \cite{yang2018hotpotqa}, use questions with subjects and objects that are explicitly named, e.g., "what color was John Wilkes Booth's hair?" \cite{kwiatkowski2019natural}, or explicitly described "What was the former band of the member of Mother
Love Bone who died just before the release of “Apple”?" \cite{yang2018hotpotqa}. The same is true of the recent conversational memory benchmarks from GoodAI and LoCoMo. As far as we can find, the "The Conversational Assistance Track Overview" (CAsT) benchmark \cite{dalton2020trec, dalton2021cast} is the only benchmark for testing IR queries generated in a conversational context that are ambiguous. All versions of the test involve retrieving from database of passages from Wikipedia, news articles, and other similar databases. The CAsT benchmark does not focus on retrieving from tabular data of conversation logs between two agents and thus does not suit our needs, but we take inspiration from this dataset in generating our own dataset, as we describe below.

\textbf{Table-based RAG.} Popular table-base retrieval datasets, like TabFact \cite{chen2019tabfact} and WikiTQ \cite{pasupat2015compositional}, involve the retrieval of information from a dataset of tabular data containing general factual information. These datasets do not test ambiguous, conversational queries and do not focus on conversation logs as we desire to do here. Current SOTA RAG systems for tabular databases write code to query and manipulate the table. For example, text-to-SQL models have shown favorable results in querying and retrieving from tabular databases, e.g., \cite{cheng2022bindingsql, ye2023datersql}. More recently, a chain-of-table (CoTable) model which chains together a series of python-like function calls shows superior performance \cite{wang2024chain}. We take inspiration from this approach in designing a model below.

\textbf{Chat-bots with Long-term Memory.} There has been a recent proliferation of chatbots with RAG both in research settings and in industry. Many open-source chatbots with memory use relatively simple semantic retrieval techniques (e.g., \cite{jang2023conversation, maharana2024evaluating, zhong2024memorybank, lee2023prompted, lu2023memochat, wu2023brief}). A growing number are using more advanced query and memory writing techniques (e.g., \cite{lu2023memochat, maharana2024evaluating, goodai_benchmark2023}). However, as far as we can find, these systems still do pure semantic retrieval or at least do not directly test the ability of the model to retrieve information based on tabular meta-data, such as queries about the time of a conversation, or the ability of the model to handle ambiguous queries. Without standard benchmarks for conversational agents which test these capabilities, it is hard to compare the performance of the various conversational agents that are being rapidly produced. We hope our work will provide a useful foundation toward resolving this issue.

\section{Dataset}\label{sec:dataset}

\textbf{The LoCoMo Dataset.} The LoCoMo dataset \cite{maharana2024evaluating} consists of 35 high-quality dialogues between two agents simulated based on the reflect and respond architecture \cite{park2023generative} using GPT4 augmented with persona prompts, a short-term memory that summarizes recent statements, and long-term memory that consists of generated observations/facts about the personas life. The generated conversations are edited by hand to ensure the conversations are consistent, non-contradictory, etc. Each conversation set consists of about 19 sessions/conversations on average, with an average of 9,209 tokens per dialogue. The LoCoMo benchmark consists of event summarization and question answering tasks, which include single-hop and multi-hop content-based questions, general knowledge questions, temporal reasoning, and an adversarial task that attempts to trick the agent into giving the wrong answer. At the time of writing, the dataset of the dialogue has been released publicly, while the questions have not been released. Nonetheless, it is clear that none of these tests focus specifically on ambiguous queries. Further, although the benchmark investigates QA tasks, the questions involved are, as far as we can tell, not designed to specifically test the ability of a model to recall portions of conversation based on meta-data retrieval, e.g., based on when the conversation occurred. The temporal reasoning task may be the closest to what we are looking for but still does not require meta-data based retrieval. Example questions from this task, such as 'How long did it take for X to finish writing the book?' \cite{maharana2024evaluating}, only require content based retrieval of conversation snippets related to book writing along with their time-stamps, which the LLM can then reason about.

\begin{figure}[t]
\includegraphics[width=.92\textwidth]{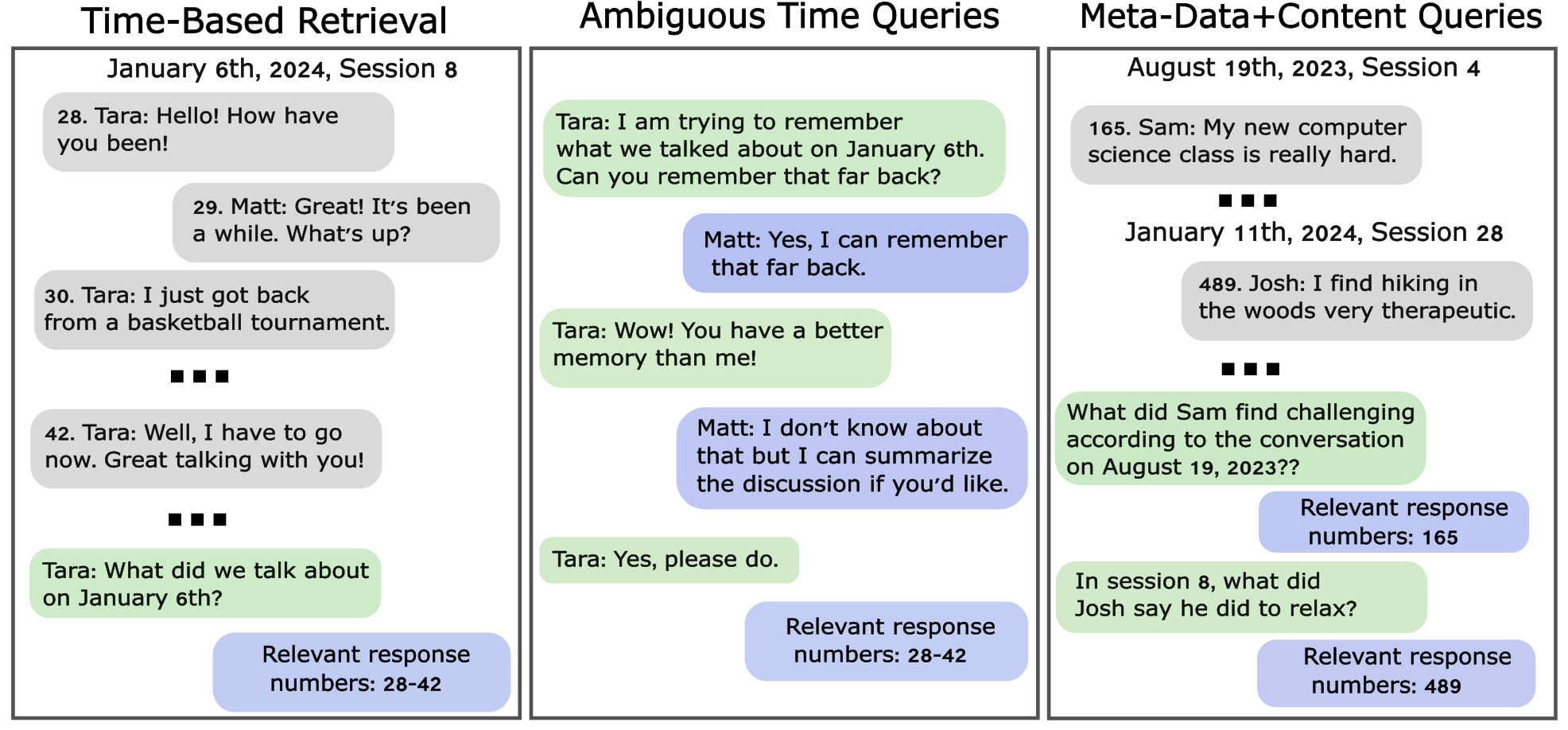}
\centering
\caption{Examples of queries for various types in our dataset.}
\end{figure}

\textbf{Modification to Locomo.} Despite the questions not yet being released, the released dialogues from LoCoMo can be used to generate new types of questions. Before generating new questions, we first made three adjustments to the original dataset. First, some of the original 35 dialogues were short, e.g.,  < 4000 tokens, which can fit directly into the context of many modern LLMs and is hence unsuitable for testing long term memory. We used only the 12 longest dialogues from the dataset. Second, we added a new conversation/session, of about 4000 tokens, to the end of each of the remaining dialogues to act as padding. We do not ask the model about this extra conversation. This ensures that the short (4k) context LLMs we are interested must use long-term memory unit that removes irrelevant recent statements from the context and replaces them with relevant statements from long-term memory. Third, the original dataset had a time and date stamp for each session/conversation. We added additional time and date stamps to each individual response where we simulate the elapsed time of a response by assuming agents generate language at a rate equal to the average number of words spoken for humans \cite{tauroza1990speech}. This augmentation allows us to test more advanced explicit time-based querying of the dataset.

\textbf{Test Set-Up} Instead of storing the desired output text/answer for each question, we store a list of the relevant responses that should be retrieved from the chat log by the IR system. This approach is similar to other IR benchmarks (e.g., \cite{dalton2021cast, dalton2020trec}) which seek to dissociate the performance of the IR/memory module from the language model that generates the answer. We believe this is a better way to isolate the performance of an IR model from that of the LLM. To measure performance, we compute the recall and F2 scores of the recalled response numbers. Recall is the proportion of relevant responses that are retrieved. F2 is the weighted harmonic mean of recall and precision, where precision is the proportion of recalled items that are relevant (see appendix \ref{app:exp_desc} for details). We use F2 instead of F1, since F2 places more emphasis on recall than precision, which better captures the desirable qualities of an IR system used with an LLM. LLMs must have relevant information in context to correctly answer a question (therefore recall of the IR system is essential). Further, it has been repeatedly found (e.g., \cite{maharana2024evaluating, goodai_benchmark2023, levy2024same}) that LLMs can ignore some but, not all, irrelevant text in context, so precision is important but less essential than recall.

Our tests assume that the agent stores the chat-log in its entirety. We believe this is feasible since chat logs, even those that may occur over the course of years, are small relative to the massive databases of texts, like wikipedia that are more commonly used in information retrieval. Nonetheless, this test can be easily adapted to slightly different forms of datasets, e.g., those that store a summary of each session instead of all responses from each session.

\textbf{Time-based Queries.} We write by hand a set of 11 different types/templates of single-hop, time-based queries that may occur in realistic conversational settings. For each question type, we use the template to automatically generate a large set of possible forms that the question can take for a given conversational set, e.g. for the question template "What did we discuss in session N?", there are N questions generated of this type per conversation set, where N is the number of sessions/conversations in the set. We then generate multiple versions of the question each with significantly different wording using several different hand crafted templates of the same question and by swapping out different spellings of numbers (e.g., writing the question both with "5th" and "fifth") for increased diversity of samples. For more detail on time based query types see appendix \ref{app:timetest}.

\textbf{Ambiguous Time-based Queries.} Next, we create ambiguous versions of the time based queries described above. We do so by creating by hand three initial conversation templates, which begin by making a comment about a conversation at a specific point in time, then proceed to talk about the conversation using demonstratives and pronouns (e.g., "it" or "that"), then finish by asking the model to summarize the conversation using a pronoun or demonstrative, such as "When did we discuss that earlier?". Multiple different wordings for the first response in the conversation are also created to generate and increase diversity of conversational templates, as well as generated conversation with different wording of numbers, times, and dates (e.g., "fifth" and "5th"), resulting an at least 6-12 different conversations per question.

\textbf{Time+Content Queries.} To better test an IR system's ability to answer more complex queries we generate multi-hop questions that require both content and meta-data based retrieval. We use GPT4 to generate the majority of the questions, by prompting it with several hand-written examples. We then edit by hand the generated questions to fix inconsistencies and ensure questions and associated response number of accurate. These questions refer to three properties of a response: 1) the speaker, 2) the date or session number, and 3) the general (non-specific) content/topic of the response, which provides some information on the topic of the response but not enough for highly accurate semantic retrieval. An example question is "What video game did Jolene mention playing with her partner on January 27th, 2023?". To get a high F2 score, it is not enough to retrieve all response related to video games, since there are dozens in the conversation logs. Nor is it enough just to retrieve all responses from January 27th or all of Jolenes responses since this will also retrieve many irrelevant responses. To maximize precision, the model must isolate the relevant response using all three features.

\section{Model}
We test and describe several simple baselines in the next section. Here we describe the more advanced information retrieval system we have designed to deal with ambiguous and meta-data-based queries in our novel dataset and benchmark tasks.

\textbf{Tabular Chat Database.} Each response consists of text as well as meta-data, such as speaker name, date, time, session number, etc. This naturally lends itself to tabular database representation, where each column in the table consists of data of a particular type (e.g., time) and each row consists of information related to a specific response. We combine a this table with a vector database by creating a 'Content' column that stores the index of the associated semantic vector in the vector database of responses. This table can then be queried for content by retrieval response from the top-k found through semantic retrieval, along with any meta-data based queries. We describe how we combine multiple queries.

\textbf{Classifying Query Type.} Some questions may require retrieving responses based on semantic retrieval, where a semantic embedder is used to create and retrieve semantically related responses from a vector database. Other questions may just require querying the table for responses with relevant meta-data. Others may require both types of query simultaneously. We use an LLM to classify whether a given query requires meta-data retrieval (by outputting yes or no) and/or semantic retrieval (yes or no). This LLM call uses few shot prompting with examples. We show empirically below that such classification can be highly accurate, and improves performance over methods which require the model to both create complex database queries and decide which type of queries to perform in the same response.

\textbf{Chain-of-Tables for Meta-Data Retrieval.} Tabular databases can be queried using standard specialized programming languages such as SQL or data-frame based python languages (e.g., pandas). One recent high performing method for querying tabular data, which outperformed SoTA text2SQL approaches, is the chain-of-table algorithm \cite{wang2024chain}. This algorithm creates a small library of functions that operates on and retrieves elements from the table. An LLM is used to call a chain of these functions in a sequence to perform advanced multi-hop queries on the table. We adapt chain of tables for meta-data queries on conversational logs. We use two functions to retrieve subsets of rows from the table:
\begin{itemize}
\item \texttt{f\_value(column\_name, [value1, value2,...])} 
\item \texttt{f\_between(column\_name, [value1, value2])}
\end{itemize}
The \texttt{f\_value} function retrieves all rows that match at least one of the listed values in the given column. The \texttt{f\_between} function retrieves all rows in between \texttt{value1} and \texttt{value2}. Although these functions are simple we find they can be used to accurately complete, in principle, all of the questions in our test set. We adapt the prompts from the original chain of table method to this custom function library. We use an LLM to write the chain using separate prompts to write 1) the function name, 2) the first argument (column name) and 3) the second argument (values) of the function. These prompts can be found in appendix \ref{app:prompts}.

\textbf{Combining Meta-Data and Semantic Retrieval.} We combine semantic and meta-data retrieval in the following way. If a query is classified as requiring both meta-data and content based retrieval, we first do a meta-data based retrieval using the chain-of-table method described above. This outputs a subset of rows from the original table. Then we to a semantic search to retrieve the top-k related related responses from the \textit{remaining} table. The intuition for ordering queries in this way is that tabular data search is a well-known and optimized problem which can be fast and deterministic while semantic search using embeddings is a stochastic and computationally expensive problem. Thus, reducing the search space by first running the cheap, deterministic queries, results in a more efficient and reliable execution.


\textbf{Query Rewrite} To deal with ambiguous queries we adapt a SoTA prompting based query rewriting method \cite{mao2023large}, which uses few-shot prompting techniques to show an LLM how to disambiguate a query given the query and preceding context. We adapt this prompt to include examples similar to our dataset, and to make sure the model does not significantly alter the query if unambiguous (see appendix \ref{app:prompts}). Although other methods exist that directly map ambiguous queries to embedding vectors (e.g., \cite{mo2023learning}), these require retraining or fine-tuning which may not be possible when accessing LLMs remotely via API.

In sum, when a query is given to our full model, it first disambiguates the query using an LLM for query rewriting. Then the query is classified as either needing only meta-data based retrieval, only semantic retrieval, or both. Finally, chain of tables and/or semantic retrieval is used to return the relevant responses and their meta-data. Pseudo-code for this algorithm is shown in appendix \ref{app:code}.

\section{Experiments}

\begin{table}
  \label{sample-table}
  \centering
  \begin{tabular}{llllllll}
    \toprule
    & \multicolumn{2}{c}{\textbf{Time Qs}}  & \multicolumn{2}{c}{\textbf{Time+Content Qs}} & \multicolumn{2}{c}{\textbf{Average}}\\
    \cmidrule(r){2-3}
    \cmidrule(r){4-5}
    \cmidrule(r){6-7}
    \textbf{Retrieval Method} & \textbf{Recall} & \textbf{F2} & \textbf{Recall} & \textbf{F2} & \textbf{Recall} & \textbf{F2}\\
    \midrule
    Semantic \text{\hspace{40pt}}(k=10) & $2.01$  & $2.32$  & $15.43$  & $5.62$  & $8.72$  & $3.97$ \\
    \text{\hspace{80pt}}(k=20) & $3.91$  & $4.21$  & $24.29$  & $5.19$  & $14.10$  & $4.70$ \\
    \text{\hspace{80pt}}(k=30) & $5.82$  & $5.89$  & $29.43$  & $4.43$  & $17.62$  & $5.16$ \\
    \midrule
    Semantic w/MetaD (k=10)  & $2.51$  & $2.90$   & $37.83$ & $13.68$  & $20.17$ & $8.29$\\
    \text{\hspace{80pt}}(k=20) & $5.02$  & $5.43$  & $51.26$  & $10.85$  & $28.14$  & $8.14$ \\
    \text{\hspace{80pt}}(k=30) & $7.47$  & $7.55$  & $56.40$  & $8.43$  & $31.93$  & $7.99$ \\
    \midrule
    CoTable+Semantic (hMistral7b) & $93.95$  & $87.67$  & $65.30$  & $22.69$   & $79.62$  & $55.18$ \\
    CoTable+Semantic (GPT3.5) & $90.47$  & $78.34$  & $90.17$  & $32.19$   & $90.32$ & $55.27$ \\
    \bottomrule
  \end{tabular}
  \caption{Recall and F2 scores for unambiguous time-based and time+content-based questions. All models use the original question to query the database.}\label{tab:main}
\end{table}

We compare our model to two baselines. First is a simple semantic retrieval system that stores a vector database of semantic vectors each representing the text of one response (semantic). Second, we test a semantic retrieval system that using a more advanced writing technique, where the meta-data is explicitly written in text then concatenated to the response text before, the theory being that this could allow pure semantic search to perform meta-data lookup by similarity matching. This combined text chunk is used to generate the semantic embedding for each response (Semantic w/ Meta Data).

We test our combined CoTable+Semantic IR system using two different LLMs with short 4k contexts for the rewrite and CoTable writing. First, we test the model with an open-source fine-tuned version of Mistral 7b \cite{jiang2023mistral} called OpenHermes (hMistral-7b) \cite{openhermes2023}, which we found performed noticeably better at writing python-like function calls during chain of table than the base Mistral model. Second, we test our model with GPT-3.5-turbo, a large model with 4k context that is relatively cheap in cost compared to its GPT4 counterparts. All of the CoTable+Semantic models retrieve k=10 values when using the semantic retrieval. All models use the same semantic embedder, the \texttt{multi-qa-mpnet-base-dot-v1} model \cite{sentencetransformer2022}, which is a high-performing, open source semantic embedder trained specifically for QA.

\begin{wrapfigure}{r}{0.4\textwidth}
  \begin{center}
    \includegraphics[width=0.38\textwidth]{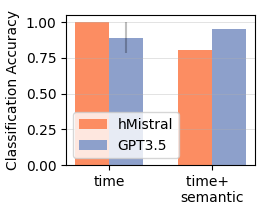}
  \end{center}
  \caption{Classification accuracy for the meta-semantic classifier on each test set.}\label{fig:classification1}
\end{wrapfigure}

\textbf{Time based Queries.} Results for time based query test (unambiguous questions) can be found in table \ref{tab:main}. The semantic retrieval model failed the time-based queries completely, which is to be expected since it uses embedding vectors that carry no information about the meta-data of the response. The semantic search with meta-data imbued embeddings performed slightly better but still performed very poorly overall, getting recall and F2 scores less than 10. Our CoTable+Semantic retrieval method performed well getting 90 recall accuracy for both LLMs and around 75-78 F2. Interestingly, GPT3.5 performed slightly worse than hMistral-7b. We find this may be due to the meta-semantic classifier being slightly less accurate with GPT-3.5 for these questions (table \ref{fig:classification1}).

\textbf{Time+Content Queries.} Results for the test with combined time and content based queries can be found in \ref{tab:main}. The basic semantic retrieval did better than it did on the pure time-based questions, which is to be expected because the embedding vectors now carry information relevant to the query, which discusses the content of the relevant response. However, because these questions are not especially specific with respect to the content of the referenced response, and because this simple IR system is insensitive to the meta-data, it still performed poorly, getting a maximum recall around $28\%$ and F2 around $5.6$. The semantic IR with meta-data imbued semantic vectors performed noticeably better than basic semantic IR getting recall closer to $56\%$ and maximum F2 around $13$. However, the CoTable+Semantic models performed much better, with the GPT3.5 model doing the best. The hMistral model obtains around $65\%$ recall accuracy and an F2 of $22.69$. The GPT-3.5 version obtains $90.17\%$ recall and around $32$ F2.

\begin{figure}[t]
\centering
\includegraphics[width=.95\textwidth]{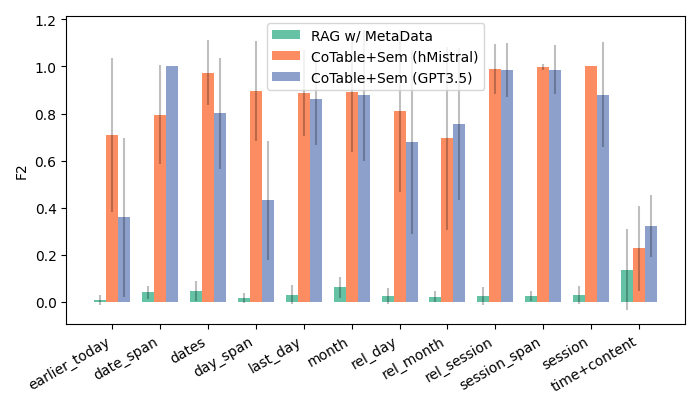}
\caption{F2 scores for each individual time-based test and the time+content based test. All models use k=10 for semantic search. Error bars show std. of recall and precision across data in each test.}
\end{figure}

\begin{table}
  \centering
  \begin{tabular}{lllllll}
    \toprule
    && \multicolumn{2}{c}{\textbf{hMistral7b}}  & \multicolumn{2}{c}{\textbf{GPT-3.5}}\\
    \cmidrule(r){3-4}
    \cmidrule(r){5-6}
    
    \textbf{Retrieval Method} & \textbf{Query Type} & \textbf{Recall} & \textbf{F2} & \textbf{Recall} & \textbf{F2}\\
    \midrule
    CoTable+Semantic & Original Qry & $2.93$  & $2.35$  & $10.62$  & $3.12$\\
    & Context+Qry & $73.51$  & $61.59$  & $77.27$  & $65.47$\\
    & Qry Rewrite & $89.43$  & $81.05$  & $83.9$  & $72.56$  \\
    \bottomrule
  \end{tabular}
  \caption{Recall and F2 scores for ambiguous time-based questions. All models use k=10 for semantic search. Models either use the original, ambiguous question to query the database (None), the query and the 3-6 preceding responses (context), or an LLM query writer prompted to disambiguate the query (Qry Write).}\label{tb:ambiguous}
\end{table}

\begin{table}
  \centering
  \begin{tabular}{llllllll}
    \toprule
    & \multicolumn{2}{c}{\textbf{Time Qs}}  & \multicolumn{2}{c}{\textbf{Time+Content Qs}} & \multicolumn{2}{c}{\textbf{Average}}\\
    \cmidrule(r){2-3}
    \cmidrule(r){4-5}
    \cmidrule(r){6-7}
    \textbf{Retrieval Method} & \textbf{Recall} & \textbf{F2} & \textbf{Recall} & \textbf{F2} & \textbf{Recall} & \textbf{F2}\\
    \midrule
    CoTable+Semantic (hMistral7b) &  & & & & & \\
    \hspace{15pt}w/ meta-semantic classification & $93.95$  & $87.67$  & $65.30$  & $22.69$   & $79.62$  & $55.18$ \\
    \hspace{15pt}w/o meta-semantic classification & $42.26$  & $33.11$  & $69.07$  & $28.29$   & $18.82$  & $8.01$ \\
    \midrule
    CoTable+Semantic (GPT3.5) &  &  & & & & \\
    \hspace{15pt}w/ meta-semantic classification & $90.47$  & $78.34$  & $90.17$  & $32.19$   & $90.32$ & $55.27$ \\
    \hspace{15pt}w/o meta-semantic classification & $89.78$  & $75.25$  & $63.10$  & $24.52$   & $76.44$  & $49.88$ \\
    \bottomrule
  \end{tabular}
  \caption{Testing the models with and without a meta-semantic classification step. Recall and F2 scores for unambiguous time-based and time+content-based questions.}\label{tab:abl1}
\end{table}

\textbf{Ambiguous Questions} Next, we tested the CoTable+Semantic IR model on ambiguous time-based questions. We did not test the other baseline models on this test since they performed poorly on the time-based questions in the unambiguous case, and thus will certainly fail in the ambiguous case. We test the CoTable+Semantic IR model first using only the original query (original query), which as expected, failed the task since it has no way to infer what is being referenced by the question. Next, we test the query rewrite method (query rewrite). This method performs similarly, though slightly worse, to the performance of the model that used the original unambiguous query. We also tested another baseline, which simply concatenated the previous sentences in the conversation chunk to the query and used the resulting conversation chunk as input to the IR system (context+query). This method performed surprisingly well achieving around 73\% recall and 61 F2 with hMistral and around 77\% recall and 65 F2 with GPT-3.5,. However, this is still significantly worse than the query rewrite. 

\textbf{Ablation Study} One key addition we made to the CoTable method was the addition of semantic retrieval and the classifier, which classifies the query as either requiring meta-data or semantic based queries or both. This was motivated by preliminary tests on CoTable which found that a common mistake in the CoTable when function calling both meta-data and the content columns, was that the model got confused about when to do content based retrieval and when not to -- models tended to perform content based search almost every query, even when content based retrieval was not appropriate. To provide empirical evidence the classifier helps mitigate this issue, we show results for a CoTable+Semantic IR model that does not use the meta-semantic classifier, but instead attempts to perform chain of table on the whole table, including the content column. The same prompts are used for this combined CoTable model, except now prompts show that there exists a content column that can be queried and have related examples. Results are shown in table \ref{tab:abl1}. We see that overall the meta-semantic classifier improves performance significantly for both Mistral and GPT-3.5. Interestingly, it helps these models in different cases: meta-semantic classification mainly helps Mistral with pure time questions, while it helps GPT-3.5 mainly in the Time+Content questions. This may be due to differences in the accuracy of the classifiers in each case (see figure \ref{fig:classification1}). In the other cases, the classifier does not seem to effect performance significantly.
\vspace{-0.4cm}
\section{Limitations}
\vspace{-0.3cm}We only tested our prompting on two models, Mistral-Hermes-7b and GPT-3.5-turbo. It is possible that other LLMs could behave differently with our prompts, though most of our prompts were derived from previous work that found these methods to be successful in other LLMS (e.g., \cite{wang2024chain}). We also did not test a variety of different semantic embedding models, and used a relatively small model (< 500 million parameters). It is possible that performance would be improved across the board with better embedding models, although it would be surprising if the relative performance of the baselines changed significantly given they all used the same retriever in our tests. Our dataset focused primarily on single-hop time-based questions, though out time+content dataset had multi-hop like questions. We also did not create ambiguous versions of the time+content based questions. Future work could focus on adding more complicated time-based questions and ambiguous versions of the time+content based questions. 
\vspace{-0.3cm}
\section{Conclusion}
\vspace{-0.3cm}Conversational AI agents may soon become common-place in our technologies, as they are integrated into virtual assistants, chat bots, and operating systems. An important capability for conversational agents is an understanding of when conversational events occurred in the past and an ability to handle ambiguous queries, which are commonplace in conversational contexts. We believe that our dataset and our system that combines chain-of-table and semantic retrieval methods makes a useful step in this direction.Our dataset provides a variety of questions that can test these ability, and our model provides strong foundation and baseline for RAG systems that can handle ambiguous and time based queries.Ultimately, long term memory systems for AI agents must combine the strengths of tabular and database-centric information retrieval systems which have been developed and optimized for decades, with the semantic flexibility and additional affordances created by embeddings and LLMs.

\bibliographystyle{plain}
\bibliography{references}

\begin{thebibliography}{10}

\bibitem{bertsch2024unlimiformer}
Amanda Bertsch, Uri Alon, Graham Neubig, and Matthew Gormley.
\newblock Unlimiformer: Long-range transformers with unlimited length input.
\newblock {\em Advances in Neural Information Processing Systems}, 36, 2024.

\bibitem{borgeaud2022improving}
Sebastian Borgeaud, Arthur Mensch, Jordan Hoffmann, Trevor Cai, Eliza Rutherford, Katie Millican, George~Bm Van Den~Driessche, Jean-Baptiste Lespiau, Bogdan Damoc, Aidan Clark, et~al.
\newblock Improving language models by retrieving from trillions of tokens.
\newblock In {\em International conference on machine learning}, pages 2206--2240. PMLR, 2022.

\bibitem{bulatov2023scaling}
Aydar Bulatov, Yuri Kuratov, Yermek Kapushev, and Mikhail~S Burtsev.
\newblock Scaling transformer to 1m tokens and beyond with rmt.
\newblock {\em arXiv preprint arXiv:2304.11062}, 2023.

\bibitem{goodai_benchmark2023}
David Castillo, Joseph Davidson, Finlay Gray, José Solorzano, and Marek Rosa.
\newblock Introducing goodai ltm benchmark, 2024.

\bibitem{chen2023extending}
Shouyuan Chen, Sherman Wong, Liangjian Chen, and Yuandong Tian.
\newblock Extending context window of large language models via positional interpolation.
\newblock {\em arXiv preprint arXiv:2306.15595}, 2023.

\bibitem{chen2019tabfact}
Wenhu Chen, Hongmin Wang, Jianshu Chen, Yunkai Zhang, Hong Wang, Shiyang Li, Xiyou Zhou, and William~Yang Wang.
\newblock Tabfact: A large-scale dataset for table-based fact verification.
\newblock {\em arXiv preprint arXiv:1909.02164}, 2019.

\bibitem{cheng2022bindingsql}
Zhoujun Cheng, Tianbao Xie, Peng Shi, Chengzu Li, Rahul Nadkarni, Yushi Hu, Caiming Xiong, Dragomir Radev, Mari Ostendorf, Luke Zettlemoyer, et~al.
\newblock Binding language models in symbolic languages.
\newblock {\em arXiv preprint arXiv:2210.02875}, 2022.

\bibitem{dalton2020trec}
Jeffrey Dalton, Chenyan Xiong, and Jamie Callan.
\newblock Trec cast 2019: The conversational assistance track overview.
\newblock {\em arXiv preprint arXiv:2003.13624}, 2020.

\bibitem{dalton2021cast}
Jeffrey Dalton, Chenyan Xiong, and Jamie Callan.
\newblock Cast 2020: The conversational assistance track overview.
\newblock In {\em In Proceedings of TREC}, 2021.

\bibitem{ding2023longnet}
Jiayu Ding, Shuming Ma, Li~Dong, Xingxing Zhang, Shaohan Huang, Wenhui Wang, Nanning Zheng, and Furu Wei.
\newblock Longnet: Scaling transformers to 1,000,000,000 tokens.
\newblock {\em arXiv preprint arXiv:2307.02486}, 2023.

\bibitem{jang2023conversation}
Jihyoung Jang, Minseong Boo, and Hyounghun Kim.
\newblock Conversation chronicles: Towards diverse temporal and relational dynamics in multi-session conversations.
\newblock {\em arXiv preprint arXiv:2310.13420}, 2023.

\bibitem{jiang2023mistral}
Albert~Q Jiang, Alexandre Sablayrolles, Arthur Mensch, Chris Bamford, Devendra~Singh Chaplot, Diego de~las Casas, Florian Bressand, Gianna Lengyel, Guillaume Lample, Lucile Saulnier, et~al.
\newblock Mistral 7b.
\newblock {\em arXiv preprint arXiv:2310.06825}, 2023.

\bibitem{kwiatkowski2019natural}
Tom Kwiatkowski, Jennimaria Palomaki, Olivia Redfield, Michael Collins, Ankur Parikh, Chris Alberti, Danielle Epstein, Illia Polosukhin, Jacob Devlin, Kenton Lee, et~al.
\newblock Natural questions: a benchmark for question answering research.
\newblock {\em Transactions of the Association for Computational Linguistics}, 7:453--466, 2019.

\bibitem{lee2023prompted}
Gibbeum Lee, Volker Hartmann, Jongho Park, Dimitris Papailiopoulos, and Kangwook Lee.
\newblock Prompted llms as chatbot modules for long open-domain conversation.
\newblock {\em arXiv preprint arXiv:2305.04533}, 2023.

\bibitem{levy2024same}
Mosh Levy, Alon Jacoby, and Yoav Goldberg.
\newblock Same task, more tokens: the impact of input length on the reasoning performance of large language models.
\newblock {\em arXiv preprint arXiv:2402.14848}, 2024.

\bibitem{lewis2020retrieval}
Patrick Lewis, Ethan Perez, Aleksandra Piktus, Fabio Petroni, Vladimir Karpukhin, Naman Goyal, Heinrich K{\"u}ttler, Mike Lewis, Wen-tau Yih, Tim Rockt{\"a}schel, et~al.
\newblock Retrieval-augmented generation for knowledge-intensive nlp tasks.
\newblock {\em Advances in Neural Information Processing Systems}, 33:9459--9474, 2020.

\bibitem{lu2023memochat}
Junru Lu, Siyu An, Mingbao Lin, Gabriele Pergola, Yulan He, Di~Yin, Xing Sun, and Yunsheng Wu.
\newblock Memochat: Tuning llms to use memos for consistent long-range open-domain conversation.
\newblock {\em arXiv preprint arXiv:2308.08239}, 2023.

\bibitem{ma2024megalodon}
Xuezhe Ma, Xiaomeng Yang, Wenhan Xiong, Beidi Chen, Lili Yu, Hao Zhang, Jonathan May, Luke Zettlemoyer, Omer Levy, and Chunting Zhou.
\newblock Megalodon: Efficient llm pretraining and inference with unlimited context length.
\newblock {\em arXiv preprint arXiv:2404.08801}, 2024.

\bibitem{maharana2024evaluating}
Adyasha Maharana, Dong-Ho Lee, Sergey Tulyakov, Mohit Bansal, Francesco Barbieri, and Yuwei Fang.
\newblock Evaluating very long-term conversational memory of llm agents.
\newblock {\em arXiv preprint arXiv:2402.17753}, 2024.

\bibitem{mao2023large}
Kelong Mao, Zhicheng Dou, Fengran Mo, Jiewen Hou, Haonan Chen, and Hongjin Qian.
\newblock Large language models know your contextual search intent: A prompting framework for conversational search.
\newblock {\em arXiv preprint arXiv:2303.06573}, 2023.

\bibitem{mo2023learning}
Fengran Mo, Jian-Yun Nie, Kaiyu Huang, Kelong Mao, Yutao Zhu, Peng Li, and Yang Liu.
\newblock Learning to relate to previous turns in conversational search.
\newblock In {\em Proceedings of the 29th ACM SIGKDD Conference on Knowledge Discovery and Data Mining}, pages 1722--1732, 2023.

\bibitem{park2023generative}
Joon~Sung Park, Joseph O'Brien, Carrie~Jun Cai, Meredith~Ringel Morris, Percy Liang, and Michael~S Bernstein.
\newblock Generative agents: Interactive simulacra of human behavior.
\newblock In {\em Proceedings of the 36th Annual ACM Symposium on User Interface Software and Technology}, pages 1--22, 2023.

\bibitem{pasupat2015compositional}
Panupong Pasupat and Percy Liang.
\newblock Compositional semantic parsing on semi-structured tables.
\newblock {\em arXiv preprint arXiv:1508.00305}, 2015.

\bibitem{sentencetransformer2022}
N.~Reimers and O.~Espejel, 2022.

\bibitem{tauroza1990speech}
Steve Tauroza and Desmond Allison.
\newblock Speech rates in british english.
\newblock {\em Applied linguistics}, 11(1):90--105, 1990.

\bibitem{openhermes2023}
teknium.
\newblock Openhermes 2.5 - mistral 7b, 2023.

\bibitem{wang2023tram}
Yuqing Wang and Yun Zhao.
\newblock Tram: Benchmarking temporal reasoning for large language models.
\newblock {\em arXiv preprint arXiv:2310.00835}, 2023.

\bibitem{wang2024chain}
Zilong Wang, Hao Zhang, Chun-Liang Li, Julian~Martin Eisenschlos, Vincent Perot, Zifeng Wang, Lesly Miculicich, Yasuhisa Fujii, Jingbo Shang, Chen-Yu Lee, et~al.
\newblock Chain-of-table: Evolving tables in the reasoning chain for table understanding.
\newblock {\em arXiv preprint arXiv:2401.04398}, 2024.

\bibitem{wu2023brief}
Tianyu Wu, Shizhu He, Jingping Liu, Siqi Sun, Kang Liu, Qing-Long Han, and Yang Tang.
\newblock A brief overview of chatgpt: The history, status quo and potential future development.
\newblock {\em IEEE/CAA Journal of Automatica Sinica}, 10(5):1122--1136, 2023.

\bibitem{yang2018hotpotqa}
Zhilin Yang, Peng Qi, Saizheng Zhang, Yoshua Bengio, William~W Cohen, Ruslan Salakhutdinov, and Christopher~D Manning.
\newblock Hotpotqa: A dataset for diverse, explainable multi-hop question answering.
\newblock {\em arXiv preprint arXiv:1809.09600}, 2018.

\bibitem{ye2023datersql}
Yunhu Ye, Binyuan Hui, Min Yang, Binhua Li, Fei Huang, and Yongbin Li.
\newblock Large language models are versatile decomposers: Decompose evidence and questions for table-based reasoning.
\newblock {\em arXiv preprint arXiv:2301.13808}, 2023.

\bibitem{zhong2024memorybank}
Wanjun Zhong, Lianghong Guo, Qiqi Gao, He~Ye, and Yanlin Wang.
\newblock Memorybank: Enhancing large language models with long-term memory.
\newblock In {\em Proceedings of the AAAI Conference on Artificial Intelligence}, volume~38, pages 19724--19731, 2024.

\end{thebibliography}


\appendix

\section{Appendix}

\subsection{Code Availability}
All code and data will be released upon publication. Our benchmark questions are included in the supplementary material. 

\subsection{Experiment Details}\label{app:exp_desc}

In the appendix experiment description we say we run LLMs and sentence embedding locally on one L40 per run, unless using GPT-3.5. Our tests for the time-based queries, ambiguous queries, and time+content queries are run in essentially the same way. The models are presented with responses in chronological order one by one, similar to how they would be presented in a real conversational scenario. In addition to the response text, models are provided the date, timestamp, and speaker of the response. Models are not provided the session number, but rather must infer what session number it is. We switch to a new session if the time gap between the current and the previous response is greater than 20 minutes. 

In all models the semantic embedder is used to store a single semantic vector per response. The Faiss library (CITE) is used to perform a flat search. The similarity measure used is cosine similarity. We use greedy sampling to generate text for chain-of-table and meta-semantic classification.

After all responses have been presented to the model, the current time is moved forward 50 minutes from the time of the last response. Then questions are presented to the model. The model is scored first on recall, which tells you the proportion of relevant response retrieved out of all relevant responses:
\begin{equation}
\text{Recall} = \frac{\text{Relevant retrieved responses}}{\text{All relevant responses}}
\end{equation}
Second, we score by the F2 score which is
\begin{equation}
\text{F2} = \frac{5 \cdot \text{precision} \cdot \text{recall}}{(4 * \text{precision}) + \text{recall}}
\end{equation}
where 
\begin{equation}
\text{Precision} = \frac{\text{Relevant retrieved responses}}{\text{All retrieved responses}}
\end{equation}
The F2 is a harmonic mean of recall and precision, where recall is weighted more heavily than precision. The motivation for F2 rather than F1 is that an LLM cannot respond to a user with a fully correct answer if it is not given all the relevant information, so recall is essential. Prior work \cite{levy2024same} has found empirically that LLMs can deal with some noise and irrelevant statements, but have been shown to perform worse on a variety of tasks when irrelevant text is present in its input, so precision should carry some weight but not as much as recall. Hence, we think F2 provides a reasonable measure of what it desired from an information retrieval system that provides input to an LLM.

Each individual test in the time-based query dataset consists of a different number of questions. To evenly weight each test when computing recall and F2 for these time based questions, we compute recall and F2 individually for each test, then average across tests.

\begin{table}
  \centering
  \begin{tabular}{cccccc}
    \toprule    
    \textbf{Test} & \textbf{Unambig. Qs} & \textbf{Unambig. Qs Full} & \textbf{Ambig. Qs} & \textbf{Ambig. Qs Full}\\
    \midrule
    earlier today & 12 & 36 & 12 & 36\\
    date span & 180 & 2160 & 180 & 1080\\
    dates & 330 & 3960 & 330 & 1980\\
    day span & 24 & 108 & 24 & 108\\
    last day & 12 & 36 & 12 & 36\\
    month & 100 & 300 & 100 & 300\\
    rel. day & 317& 938 & 304 & 912\\
    rel. month& 100 & 264 & 100 & 276\\
    rel session & 330 & 1014 & 330 & 1002\\
    session span & 258 & 1032 & 258 & 1032\\
    session & 294 & 1764 & 294 & 1764\\
    time+content & 177 & - & - & -\\
    total   &2134 & 11612& 1944 & 8526\\
    \bottomrule
  \end{tabular}
  \caption{Number of questions per test. We show the number questions for unambiguous (unambig.) and ambiguous (ambig.) questions, both when we sum together every question and when we sum together every version of every question (full). }\label{fig:number_qs_dataset}
\end{table}

\subsection{Time Test Question Descriptions}\label{app:timetest}
Here we describe each type of question that shows up in the time-based query tests.
\begin{itemize}
    \item The \texttt{earlier\_today} test poses variations on the question "what did we discuss earlier today?".
    \item The \texttt{date\_span} test poses variations of the question "what did we discuss between DATE1 and DATE2?"
    \item The \texttt{dates} test poses variation of the question "what did we discuss on DATE?"
    \item The \texttt{day\_span} test poses variations of two questions "what did we talk about over the last three days?" and "what did we talked about over the last week?".
    \item The \texttt{last\_day} test poses variations of the question "what did we discuss last DAY NAME?", where DAY NAME can be Monday, Tuesday, etc.
    \item The \texttt{month} test poses variations of the question " what did we discuss in MONTH, YEAR?"
    \item The \texttt{rel\_day} test poses variations of the question "what did we discuss N days ago?"
    \item The \texttt{rel\_month} test poses variations of the question "what did we discuss N months ago?"
    \item The \texttt{rel\_session} test poses variations of the question "what did we discuss N sessions ago?"
    \item The \texttt{session\_span} test poses variations of the question 'what did we talk about between session/conversation/discussion S1 and S2?'
    \item The \texttt{session} test poses variations of the question "What did we discuss in session/discussion/conversation N?"
\end{itemize}

\subsection{Templates for Ambiguous Queries}
The following are the templates used to create the surrounding context for the ambiguous time based questions. Here \texttt{X} is replaced with one of the temporal stamps noted in the previous section, e.g. "N session ago" or "on MONTH YEAR". These base templates a varied slightly as needed for each test, and further variations where made using different spellings of numerals.

\textbf{Template 1}

\texttt{speaker\_a: I see in my calendar that we talked X.}

\texttt{speaker\_b: Yes! we did talk then. I always enjoy our chats.}

\texttt{speaker\_a: I enjoy them too! Can you summarize what we discussed?}
\newline

\textbf{Template 2}

\texttt{speaker\_a: I am try to remember what we talked about X. Can you remember that far back?}

\texttt{speaker\_b: Yes, I can remember that far back.}

\texttt{speaker\_a: Wow! You have a better memory than me!}

\texttt{speaker\_b: I don't know about that! But I can summarize our discussion for you if you'd like.}

\texttt{speaker\_a: Yes, please do.}
\newline

\textbf{Template 3}

\texttt{speaker\_a: I remember X we had several discussions.}

\texttt{speaker\_b: Yes, we did.}

\texttt{speaker\_a: But I cannot quite remember what we discussed.}

\texttt{speaker\_b: Would you like me to tell you?}

\texttt{speaker\_a: Yes, could you describe, in as much detail as you can, the content of those conversations?}

\subsection{Prompting for Context Chunk to Chain-of-Table}
In our ambiguous query test we found that of the methods we tried, disambiguating the query through a prompt based rewrite method worked the best. However, we found a simple baseline where we simply inject a chunk of previous context (i.e., the query plus the preceding 2-4 statements) significantly improves performance without changing our original chain of table prompts. To see if we could improve performance further, we augmented our chain of table prompt with examples that included chunks of the conversational context rather than just a single query/statement. Results are shown in table \ref{fig:contextprompt}. We found this did not improve this method over the original prompt.

\begin{table}
  \centering
  \begin{tabular}{lllllll}
    \toprule
    &&& \multicolumn{2}{c}{\textbf{hMistral7b}}\\
    \cmidrule(r){4-5}
    \cmidrule(r){6-7}
    
    \textbf{Retrieval Method} & \textbf{Query Type} & \textbf{Prompt Type} & \textbf{Recall} & \textbf{F2}\\
    \midrule
    CoTable+Semantic & Context+Qry & Original CoTb Prompt & $73.51$  & $61.59$\\
    & Context+Qry & Context CoTb Prompt & $74.56$  & $52.19$ \\
    \bottomrule
  \end{tabular}
  \caption{Recall and F2 scores for ambiguous time-based questions. }\label{fig:contextprompt}
\end{table}

\subsection{Algorithm Description}\label{app:code}

\begin{algorithm}
\caption{Chain-of-Table w/ Semantic Retrieval}\label{alg:cap}
\begin{algorithmic}
\State \textbf{Data: } $(T,Q)$ \Comment{Table and query}
\State \textbf{Result: } $T_{new}$ \Comment{New Table, with only relevant rows}
\Function{CoTable+Semantic}{$T,Q$}
\State $T_{new} = T$
\State chain = []
\State do\_meta, do\_semantic = classify(Q) \Comment{classify query} 
\If{do\_meta}
\While{$f \neq$ <END>} \Comment{If q is referencing meta-data, do chain of table} 
\State f = get\_function($T_{new}, Q$, chain)
\State arg1 = get\_arg1(f, $T_{new}, Q$, chain)
\State arg2 = get\_arg2(f, arg1, $T_{new}, Q$, chain)
\State $T_{new}$ = apply\_func(f, arg1, arg2, $T_{new}$)
\State chain.append([f, arg1, arg2])
\EndWhile
\EndIf
\If{do\_semantic}: \Comment{If referencing content, do a semantic search} 
\State $T_{new}$ = semantic\_search($T_{new}, Q$)
\EndIf
\State \textbf{return} $T_{new}$
\EndFunction
\end{algorithmic}
\end{algorithm}

\newpage
\subsection{Prompts}\label{app:prompts}
Below are descriptions of prompts used in our meta-semantic classifier, chain-of-table writer, and query rewriter.

\subsubsection{Meta-Classify Prompt}
\colorbox{prompt_gray}{\begin{minipage}{.98\textwidth}
\texttt{We have a table that stores a chat log of responses between two speakers in a table format. We store 1) semantic embeddings of the responses so we can search for responses with similar content, and 2) meta-data such as  the times, dates, session number, and response number for each response. Each row stores information about one response. The columns in the table are:}
\newline

\texttt{/*}

\texttt{Response\_Index | Session\_Index | Speaker | Day\_Name | Week | Date | Time}

\texttt{*/}
\newline

\texttt{We need to decide if the user's query is referring to the meta-data of the chat log or not. You will be presented with a query from the user. Answer whether the query is referring to the meta-data of previous dialogues, such as the time, date, session or response number of previous portions of discussions. If the query is referring to meta-data, respond 'y' for yes. If the query is not referring to meta-data, respond 'n' for no. Only output a 'n' or 'y' character and nothing else. For example,}
\newline

\texttt{[EXAMPLES]}
\newline

\texttt{Query: [QUERY]}

\texttt{Output: [OUTPUT]}
\end{minipage}}

\newpage
\subsubsection{Semantic-Classify Prompt}
\colorbox{prompt_gray}{\begin{minipage}{.98\textwidth}
\texttt{We have a table that stores a chat log of responses between two speakers in a table format. We store 1) semantic embeddings of the responses so we can search for responses with similar content, and 2) meta-data such as  the times, dates, session number, and response number for each response. Each row stores information about one response. The columns in the table are:}
\newline

\texttt{/*}

\texttt{Response\_Index | Session\_Index | Speaker | Day\_Name | Week | Date | Time}

\texttt{*/}
\newline

\texttt{We need to decide if the user's query is referring to some specific topic or content, or if the query is only referring to meta-data or non-specific content. You will be presented with a query from the user. Answer whether the query is on some specific topic or content. If the query is referring specific topic or content, respond 'y' for yes. If the query is not referring to a specific topic or content, respond 'n' for no. Only output a 'n', or 'y' character and nothing else. For example,}
\newline

\texttt{[EXAMPLES]}
\newline

\texttt{Query: [QUERY]}

\texttt{Output: [OUTPUT]}
\end{minipage}}

\newpage
\subsubsection{Function Write Prompt}
The function write prompt writes the function name at each step of the chain of tables algorithm.

\colorbox{prompt_gray}{\begin{minipage}{.98\textwidth}
\texttt{We have a table that stores a log of responses between two speakers in a table format. Information about speakers, response times, indexes, sessions, and dates are stored in the table, where each row stores information about one response. There are functions that allow us to isolate rows in the table. The columns in the table are:}
\newline

\texttt{/*}

\texttt{Response\_Index | Session\_Index | Speaker | Day\_Name | Week | Date | Time}

\texttt{*/}
\newline

\texttt{If the table only needs rows that have a certain value in a certain column to answer the question, we use 'f\_value(column\_name, [row\_value1, row\_value2, ...])' to select these rows. For example,}
\newline

\texttt{[EXAMPLES]}
\newline

\texttt{If the table only needs rows with that have values within a certain range, to answer the question we use 'f\_between(column\_name, [min\_value, max\_value])' to select these rows. For example,}
\newline

\texttt{[EXAMPLES]}
\newline

\texttt{If do not need to call a function, write <END>.}
\newline

\texttt{Here are examples of using the operations to respond to the query. Be sure to end the operation chain after a few function calls with <END>. Do not repeat function calls.}
\newline

\texttt{[EXAMPLES]}
\newline

\texttt{Current Date:[DATE] Current Time:[TIME] Current Session:[SESSION NUM]}

\texttt{Query: [QUERY]}

\texttt{Function Chain: [FUNCTION CHAIN SO FAR] -> [OUTPUT]}
\end{minipage}}

\newpage
\subsubsection{Argument 1 Write Prompt for \texttt{f\_value}}
The first argument the f\_value function is written by an LLM using the following prompt.

\colorbox{prompt_gray}{\begin{minipage}{.98\textwidth}
\texttt{We have a table that stores a log of responses between two speakers in a table format. Information about response speaker names, and dates are stored in the table, where each row stores information about one response. There are functions that allow us to isolate rows in the table. The columns in the table are:}
\newline

\texttt{/*}

\texttt{Response\_Index | Session\_Index | Speaker | Day\_Name | Week | Date | Time}

\texttt{*/}
\newline

\texttt{If the table only needs rows that have a certain value in a certain column to answer the question, we use 'f\_value(column\_name, [row\_value1, row\_value2, ...])' to select these rows. For example,}
\newline

\texttt{[EXAMPLES]}
\newline

\texttt{Here are examples of chaining together multiple function calls. Each chain of function calls is ended when the model outputs <END>. For example,}
\newline

\texttt{[EXAMPLES]}
\newline

\texttt{Finish the following function chain. Do not write any extra text. Only output the first argument of the 'f\_value' function.}
\newline

\texttt{Current Date:[DATE] Current Time:[TIME] Current Session:[SESSION NUM]}
\texttt{Query: [QUERY]}

\texttt{Function Chain: [FUNCTION CHAIN SO FAR] -> [OUTPUT]}
\end{minipage}}

\newpage
\subsubsection{Argument 1 Write Prompt for \texttt{f\_between}}
The first argument the f\_value function is written by an LLM using the following prompt.

\colorbox{prompt_gray}{\begin{minipage}{.98\textwidth}
\texttt{We have a table that stores a log of responses between two speakers in a table format. Information about response speaker names, and dates are stored in the table, where each row stores information about one response. There are functions that allow us to isolate rows in the table. The columns in the table are:}
\newline

\texttt{/*}

\texttt{Response\_Index | Session\_Index | Speaker | Day\_Name | Week | Date | Time}

\texttt{*/}
\newline

\texttt{If the table needs rows between two values in a certain column, we use 'f\_between(column\_name, [min\_value, max\_value])' to select these rows. For example,}
\newline

\texttt{[EXAMPLES]}
\newline

\texttt{Here are examples of chaining together multiple function calls. Each chain of function calls is ended when the model outputs <END>. For example,}
\newline

\texttt{[EXAMPLES]}
\newline

\texttt{Finish the following function chain. Do not write any extra text. Only output the first argument of the 'f\_between' function.}
\newline

\texttt{Current Date:[DATE] Current Time:[TIME] Current Session:[SESSION NUM]}

\texttt{Query: [QUERY]}

\texttt{Function Chain: [FUNCTION CHAIN SO FAR] -> [OUTPUT]}
\end{minipage}}

\newpage
\subsubsection{Argument 2 Write Prompt for \texttt{f\_value}}
The first argument the f\_value function is written by an LLM using the following prompt.

\colorbox{prompt_gray}{\begin{minipage}{.98\textwidth}
\texttt{We have a table that stores a log of responses between two speakers in a table format. Information about response speaker names, and dates are stored in the table, where each row stores information about one response. There are functions that allow us to isolate rows in the table. The columns in the table are:}
\newline

\texttt{/*}

\texttt{Response\_Index | Session\_Index | Speaker | Day\_Name | Week | Date | Time}

\texttt{*/}
\newline

\texttt{We are in the process of isolating the rows relevant to the query through a series of function calls. Our next step is to decide which rows to isolate using the f\_value function. The f\_value function returns all rows that have a value equal to at least one of the given values, i.e., f\_value(column\_name, [value\_1, value\_2,...]). Here are examples of how to fill in the values for the second argument:}
\newline

\texttt{[EXAMPLES]}
\newline

\texttt{Fill in the remaining argument of the f\_value function. Do not explain your answer, just provide the missing values.}
\newline
\texttt{Current Date:[DATE] Current Time:[TIME] Current Session:[SESSION NUM]}

\texttt{Query: [QUERY]}

\texttt{Function Chain: [FUNCTION CHAIN SO FAR]}
\texttt{Answer: [OUTPUT]}
\end{minipage}}
\newpage

\subsubsection{Argument 2 Write Prompt for \texttt{f\_between}}
The first argument the f\_value function is written by an LLM using the following prompt.

\colorbox{prompt_gray}{\begin{minipage}{.98\textwidth}
\texttt{We have a table that stores a log of responses between two speakers in a table format. Information about response speaker names, and dates are stored in the table, where each row stores information about one response. There are functions that allow us to isolate rows in the table. The columns in the table are:}
\newline

\texttt{/*}

\texttt{Response\_Index | Session\_Index | Speaker | Day\_Name | Week | Date | Time}

\texttt{*/}
\newline

\texttt{We are in the process of isolating the rows relevant to the query through a series of function calls. Our next step is to decide which rows to isolate using the f\_between function. The f\_between function returns all rows that have a value between a given minimum and maximum value for a given column, i.e., f\_between(column\_name, [min\_value, max\_value]). Here are examples of how to fill in the values for the second argument:}
\newline

\texttt{[EXAMPLES]}
\newline

\texttt{Fill in the remaining argument of the f\_between function. Do not explain your answer, just provide the missing values.}

\texttt{Current Date:[DATE] Current Time:[TIME] Current Session:[SESSION NUM]}

\texttt{Query: [QUERY]}

\texttt{Function Chain: [FUNCTION CHAIN SO FAR]}

\texttt{Answer: [OUTPUT]}
\end{minipage}}

\newpage
\subsubsection{Query Rewrite Prompt}
\colorbox{prompt_gray}{\begin{minipage}{.98\textwidth}
\texttt{Please help reformulate the question into a rewrite that can fully express the user's information needs without the need of context and while removing irrelevant sentences. Here are several example dialogues, where each turn contains a portion of dialogue and a rewrite.}
\newline

\texttt{[EXAMPLES]}
\newline

\texttt{If the query is not ambiguous or if there are no preceding sentences, just repeat the question. For example,}
\newline

\texttt{[EXAMPLES]}
\newline

\texttt{Query: [QUERY]}

\texttt{Rewrite: [OUTPUT]}
\end{minipage}}

\end{document}